\documentclass[conference]{IEEEtran}

\makeatletter
\def\ps@headings{%
\def\@oddhead{\mbox{}\scriptsize\rightmark \hfil \thepage}%
\def\@evenhead{\scriptsize\thepage \hfil \leftmark\mbox{}}%
\def\@oddfoot{}%
\def\@evenfoot{}}
\makeatother
\usepackage{algorithm}
\usepackage{algorithmicx}
\usepackage{algpseudocode}
\usepackage{amsmath}
\usepackage{graphicx}
\usepackage{subfigure}
\usepackage{times}
\usepackage[hyphens]{url}
\usepackage{color}
\usepackage{multirow}
\usepackage{textcomp}
\usepackage{threeparttable}
\usepackage{supertabular}
\usepackage{amssymb}
\usepackage{amsfonts}
\usepackage{booktabs}
\usepackage{epstopdf}
\usepackage{xspace}
\usepackage{enumitem}
\usepackage{cite}
\usepackage{fancyhdr}

\newcommand{\BotTriNet}{{\sc BotTriNet}\xspace}
\newcommand{\Triple}{{\sc Triple}\xspace}

\newcommand{\Cresci}{{\sc Cresci2017}\xspace}

\newcommand{\SBot}{{\small\sf SocialBots}\xspace}
\newcommand{\TBot}{{\small\sf TraditionalBots}\xspace}
\newcommand{\FBot}{{\small\sf FakeFollowers}\xspace}
\renewcommand{\paragraph}[1]{\smallskip\noindent {\bf #1}}

\begin{document}

%----------------------------------------------------------------------------
\title{BotTriNet: A Unified and Efficient Embedding for Social Bots Detection via Metric Learning}

\author{\IEEEauthorblockN{1\textsuperscript{st} Jun Wu$^*$}
\IEEEauthorblockA{
%\textit{College of Computing} \\
\textit{Georgia Institute of Technology}\\
Atlanta, United States \\
jwu772@gatech.edu}
\and
\IEEEauthorblockN{2\textsuperscript{nd} Xuesong Ye}
\IEEEauthorblockA{
%\textit{Department of Engineering and Technology} \\
\textit{Trine University}\\
Phoenix, United States \\
xye221@my.trine.edu}
\and
\IEEEauthorblockN{3\textsuperscript{rd} Yanyuet Man}
\IEEEauthorblockA{
%\textit{Department of Engineering and Technology} \\
\textit{Tencent AI Lab}\\
Shenzhen, China \\
manyanyuetnk@gmail.com}
}

\maketitle

\thispagestyle{fancy}
\fancyhead{}
\lhead{}
\lfoot{979-8-3503-3698-6/23/\$31.00 ©2023 IEEE \hfill}
\cfoot{}
\rfoot{}

\begin{abstract}
%A persistently popular topic in online social networks is the rapid and accurate discovery of bot accounts to prevent their invasion and harassment of genuine users. We propose a unified embedding framework called \BotTriNet, which utilizes textual content posted by accounts for bot detection based on the assumption that contexts naturally reveal account personalities and habits. Content is abundant and valuable if the system efficiently extracts bot-related information using embedding techniques.
%Beyond the general embedding framework that generates word, sentence, and account embeddings, we design a triplet network to tune the raw embeddings (produced by traditional natural language processing techniques) for better classification performance. We evaluate detection accuracy and F1score on a real-world dataset \Cresci, comprising three bot account categories and five bot sample sets. Our system achieves the highest average accuracy of 98.34\% and F1score of 97.99\% on two content-intensive bot sets, outperforming previous work and becoming state-of-the-art. It also makes a breakthrough on four content-less bot sets, with an average accuracy improvement of 11.52\% and an average f1score increase of 16.70\%.

The rapid and accurate identification of bot accounts in online social networks is an ongoing challenge. In this paper, we propose \BotTriNet, a unified embedding framework that leverages the textual content posted by accounts to detect bots. Our approach is based on the premise that account personalities and habits can be revealed through their contextual content. To achieve this, we designed a triplet network that refines raw embeddings using metric learning techniques. The \BotTriNet framework produces word, sentence, and account embeddings, which we evaluate on a real-world dataset, \Cresci, consisting of three bot account categories and five bot sample sets. Our approach achieves state-of-the-art performance on two content-intensive bot sets, with an average accuracy of 98.34\% and f1score of 97.99\%. Moreover, our method makes a significant breakthrough on four content-less bot sets, with an average accuracy improvement of 11.52\% and an average f1score increase of 16.70\%. Our contribution is twofold: First, we propose a unified and effective framework that combines various embeddings for bot detection. Second, we demonstrate that metric learning techniques can be applied in this context to refine raw embeddings and improve classification performance. Our approach outperforms prior works and sets a new standard for bot detection in social networks.
\end{abstract}


\section{Introduction}
\label{sec:intro}

\subsection{Background and Motivation}
Bot accounts are a persistent issue across popular internet applications and services such as Twitter, Facebook, Reddit, and Instagram, posing significant security risks. Bots engage in various malicious activities, including disseminating fake news \cite{liu2018early}, spamming \cite{rodrigues2022real}, spreading rumors \cite{huang2022social}, and promoting misinformation \cite{himelein2021bots}. However, the proliferation of text generation techniques and their applications have made distinguishing bots from humans increasingly challenging for ordinary users. For example, Yizhe \emph{et al.} \cite{zhang2019dialogpt} designed a conversational response generation technique. Rik \emph{et al.} \cite{koncel2019text} could generate coherent multi-sentence texts from the knowledge graph. Ziheng \emph{et al.} \cite{chen2022grease} and \cite{chen2023dark} investigated counterfactual explanations for the recommendation system. These factors significantly raise the difficulty and cost for companies defending against social bot attacks. They can generate harmful content and activities such as posting, digging, and commenting, increasing companies' maintenance and risk control costs while negatively impacting normal users' well-being and finances. Since most account behaviors are in textual format, an efficient content-based detection system can mitigate bots' harmful effects.

\subsection{Social Bots Detection}
\label{subsec:related}
Social bot accounts have become a growing concern for users and researchers. To address this issue, researchers have proposed various approaches for bot detection. According to the information used, these approaches can be categorized into two types: \emph{side information approach} and \emph{content approach}. The \emph{side information approach} commonly utilizes accounts' behavior and profile features. These data do not directly reveal bots' attacking intentions but serve as hidden and correlated metrics reflecting bots' homogeneity. For example, our previous work BotShape \cite{wu2023botshape} used the tweeting logs of accounts to generate behavioral sequences and patterns as significant features for bot detection. Kai-Cheng \emph{et al.}  \cite{yang2020scalable} chose the count of follower and its growth rate as core features to detect bots. Hrushikesh \emph{et al.}  \cite{shukla2021enhanced} used a comprehensive account features, e.g., location, tweet count, avatar and applied ensemble classifier to predict bot accounts. The \emph{content approach} focuses on capturing text semantics from accounts' historical postings and using classifiers to separate bots from genuine users. This approach assumes that bots' text information tends to contain attack intentions, making their semantics distinct from genuine users. For example, Feng \emph{et al.} \cite{wei2019twitter} employed bidirectional LSTM neural networks and word embeddings to enhance Twitter bot detection accuracy. Atheer \emph{et al.} \cite{alhassun2022combined} introduced a deep-learning framework that integrates textual and metadata features to improve spam account detection on Twitter. BIC \cite{lei2022bic} leveraged text-graph interaction and semantic consistency to enhance Twitter bot detection.

\subsection{Metric Learning and Triplet Network} Metric learning focuses on learning a representation between data points to improve performance in machine learning tasks such as classification, clustering, and ranking. It succeeds in many application fields thanks to its power to actively adjust the distances between classes. Eric \emph{et al.} \cite{xing2002distance} proposed a novel technique for learning a distance metric suitable for clustering tasks with side information, leading to enhanced clustering quality across various applications. Sergey \emph{et al.} \cite{zagoruyko2015learning} utilized a Siamese Network architecture for metric learning, effectively comparing image patches for a wide range of image-matching tasks. \emph{Triplet Network} is a popular method in Metric Learning that learns a similarity metric by minimizing intra-class distance and maximizing inter-class distance, mainly used and thriving in the face recognition field. Florian \emph{et al.} \cite{schroff2015facenet} introduced FaceNet, which leverages triplet loss to create a unified embedding serving multiple purposes, such as face recognition, verification, and clustering. Zhuoyi \emph{et al.} \cite{wang2020adaptive} and \cite{gao2019sim} designed a general metric-learning solution across multimodal data to deal with open-world problems. Jun \emph{et al.} \cite{wu2023fineehr} refined the textual representations of electronic health records through Siamese Network and made a better accuracy performance for mortality prediction of patients. 

\subsection{Contributions} 
Our research aims to design a unified and efficient embedding framework for social bot detection. We used the historical content posted by accounts to profile an account and applied metric learning techniques to refine content embedding and improve downstream prediction performance. To summarize, this paper makes the following contributions: 

\begin{itemize} 
\item We design a unified and effective embedding framework called \BotTriNet to produce word-level, content-level, and account-level embeddings from data. It takes the historical contents of accounts as input and detects bots based on multi-level embeddings.

\item \BotTriNet is the first practice applying metric learning techniques in the bot detection field. It represents raw content embeddings for better account profiling, significantly improving the separability between bots and genuine accounts in feature space.

\item Our method achieves an average accuracy of 95.80\% and an f1score of 94.36\% on the real-world dataset \Cresci, surpassing the performance of prior works such as \cite{davis2016botornot}, \cite{ahmed2013generic}, \cite{cresci2016dna}, and \cite{wei2019twitter}. 

\item Especially for content-less account categories, \BotTriNet can break the accuracy bottleneck of the baseline approach, gaining an average accuracy improvement of 11.52\% and an average f1score improvement of  16.70\%. 

\end{itemize}

\section{Dataset}
\label{sec:dataset}

Cresci {\em et al.} \cite{cresci2017paradigm} published a real-world dataset of Twitter (called \Cresci). Due to its authenticity and richness, have become one of the most important datasets for subsequent bot detection-related work. It contains four types of accounts, including genuine users, social bots, traditional bots, and fake followers. \emph{Genuine users} refer to accounts controlled by real human beings. The latter three account types are all not real persons, controlled by black and grey industries. \emph{Social bots} mimics human-like behavior and interactions and spreads specific content to influence opinions or manipulate discussions. \emph{Traditional bots} are automated accounts focused on spamming or advertising, with less emphasis on imitating human behavior. \emph{Fake followers} are inactive accounts created to inflate other users' follower counts, often for boosting credibility or promoting popularity.

\section{Design}
We present \BotTriNet, a unified embedding framework for social bots detection. It takes textual content posted by accounts as input data and predicts whether the account is a bot. 

\BotTriNet designs multiple-levels embeddings framework to presents content and account profiling in semantic space properly: (i) word embedding: reflecting words to vectors, (ii) content embedding: reflecting contents posted by users like tweets to vectors via its words, (iii) account embedding: profile an account according to its historical sentence embedding. The higher-level embedding is aggregated from the lower-level embeddings of the same entity, which is called pooling.

Our system not only employs the pooling technique to aggregate low-level embeddings into high-level ones, but also prioritizes optimizing the low-level embeddings using a metric learning approach before the pooling operation. Simply applying text embedding and pooling technique, and treating bot detection as a text classification task, could not  perform accurate in \Cresci, especially on some bot categories with less content like \TBot and \FBot. To tackle this problem, \BotTriNet designs a triplet network to represent raw content embedding to new feature space. We abstract a concept called \emph{Account Anchor} and use the \emph{Triplet Learning} to adjust the distance among embeddings in various anchors. Experiment results ~\ref{sec:evaluation} show this method vastly improved the classification accuracy. We will introduce each module of \BotTriNet and discuss how to design and implement our system.

\begin{figure*}[htb]
\centering
\includegraphics[width=\linewidth]{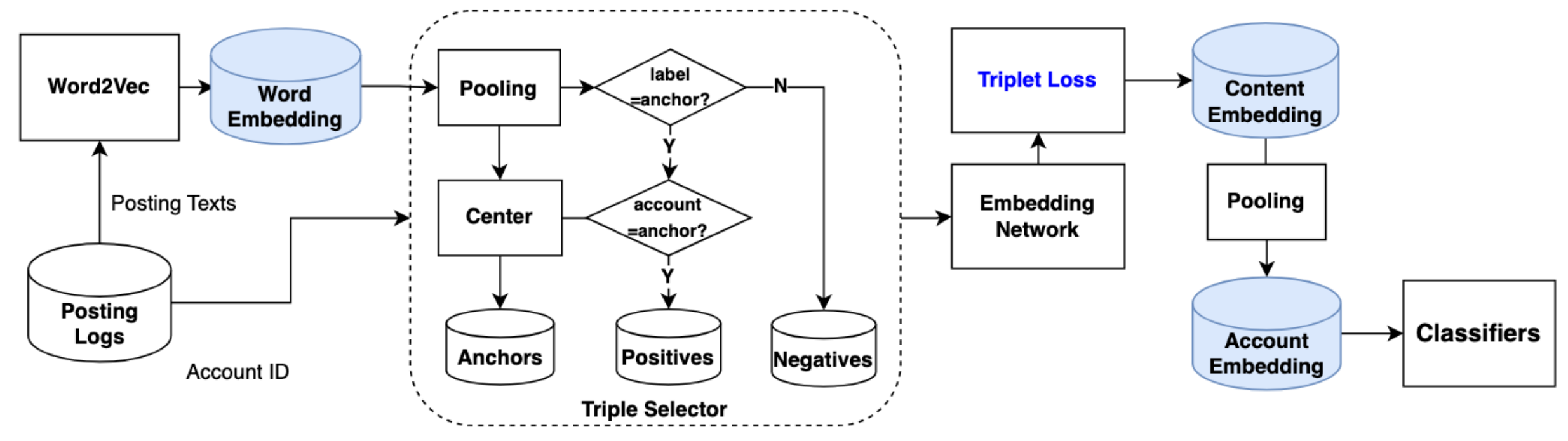}
\vspace{-6pt}
\caption{\BotTriNet architecture.}
\label{fig:arch}
\end{figure*}
\label{fig:design}

\subsection{Word Representation and Content Embeddings}
% ok
Word representation is a technique NLP uses to convert words into dense, low-dimensional vectors that capture their semantic meaning. By mapping words to a lower-dimensional space where similar words are closer, performing NLP tasks such as sentiment analysis, machine translation, and text classification is easier. Popular algorithms for word embedding include Word2Vec \cite{mikolov2013efficient}, GloVe \cite{pennington2014glove}, and FastText \cite{bojanowski2017enriching}. 

Word representation mainly has two steps: \emph{word tokenization} and \emph{word embedding generation}. \BotTriNet takes the tweeting texts in \Cresci2017 as the corpus for both of them. Each tweet is a textual sentence, and the system splits each sentence into continuous tokens via spaces and symbols. Tokens of the same type are changed into unified symbols for integrating tokens of the same type. For example, tokens starting with ``http:''  are transformed to ``[URL]''; grams starting with ``@'' are changed to ``[Others]''; grams starting with ``\#'' are changed to ``[Topic]'' and so on. Next, \BotTriNet uses Word2Vec to generate domain-specific word embeddings from a large corpus of text data in \Cresci. We selected Word2Vec because it performed best (bot classification accuracy) in our experiments. We had considered using pre-trained word embeddings; however, they only performed slightly better on \FBot and \TBot, while Word2Vec greatly improved performance on \SBot. Our system computes the average of all word embeddings as the default embedding to represent the new words that appeared in prediction tasks. 

We stipulate that \emph{content embedding} represents a text (such as a tweet) posted by an account, which has the same dimension size as word embedding. Each dimension is the average of feature values in that dimension of all words in that sentence, and this method is called \emph{average pooling}. Pooling is a technique that reduces feature map size while preserving important information. Nal \emph{et al.} \cite{kalchbrenner2014convolutional} used word embeddings as input and applied average pooling to obtain a fixed-size sentence embedding. The model achieved state-of-the-art results on several sentence classification tasks. \BotTriNet also employs pooling to produce higher-level embeddings. 

\subsection{Triplet Learning}
\begin{figure}[htb]
\centering
\includegraphics[width=\linewidth]{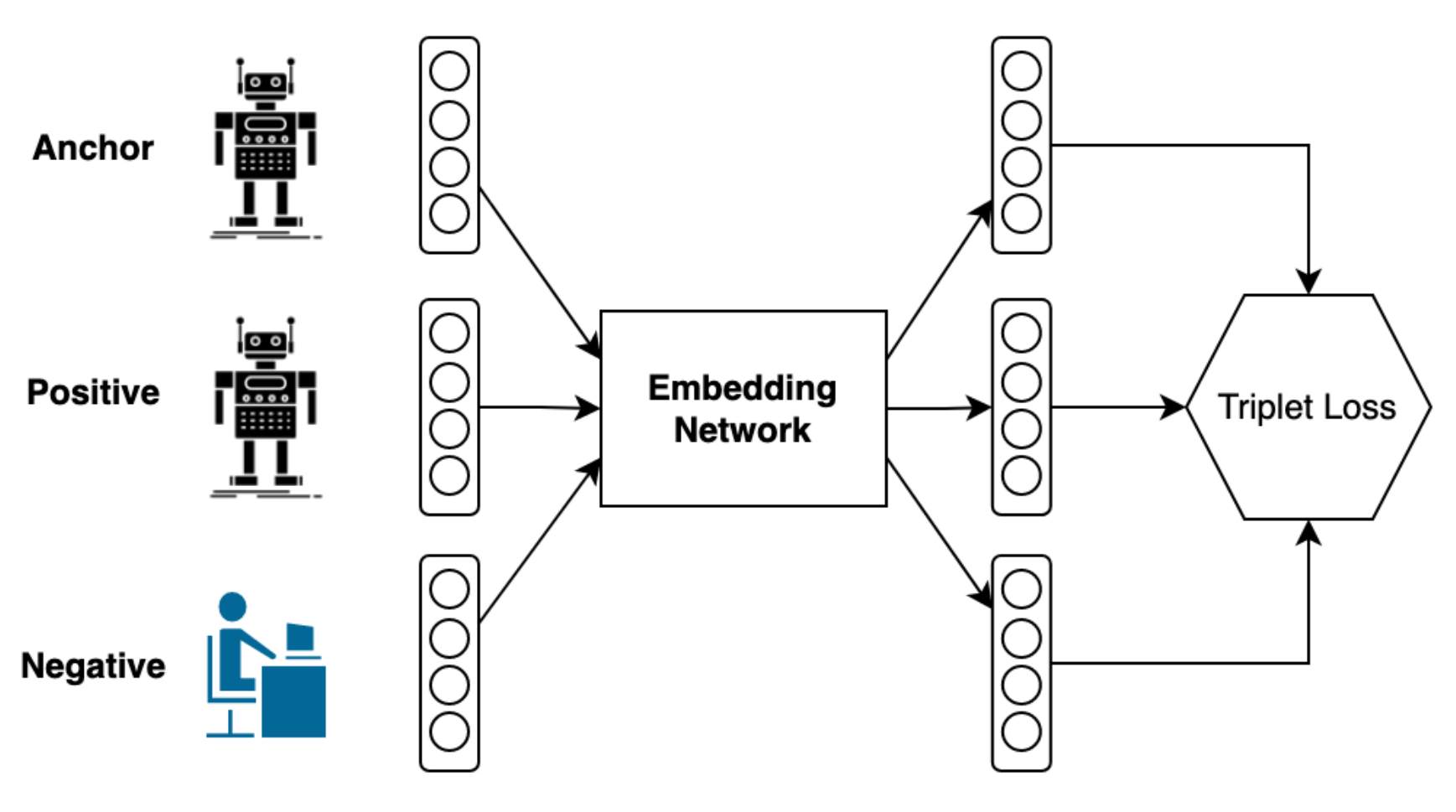}
\vspace{-6pt}
\caption{Triplet Learning}
\label{fig:arch}
\end{figure}
\label{fig:design}
Before generating account embedding via pooling, \BotTriNet refines the content embeddings by the Triplet Learning approach. We were inspired by the work of Sentence BERT \cite{reimers2019sentence}, which points out that the CLS token of the BERT model is not an ideal representation of sentences, but adjusting raw embeddings using Siamese and Triplet Learning can significantly enhance the performance of downstream text tasks. We chose Triplet Learning to refine the raw content embeddings because social bots are naturally similar to Triplet Learning's definition, which contains a core concept called Anchor. In object recognition, the anchor is a collective concept like a cat or human. Similarly, each account is also an anchor in social media, and its behavior reflects account activities, opinions, and habits. 

Specifically, \BotTriNet designed a specific triplet learning algorithm for social bot detection by modifying its original modeling process to adapt to the social account scenario and the social bot detection task. We have two objectives: (i) maximize the distance between bots and genuine users, and (ii) improve the classification accuracy of account embeddings. We split the whole learning process into three modules: \emph{Embedding Network}, \emph{Triplet Loss} and \emph{Triple Selector}. The Embedding Network, actually a neural network, undertakes to represent the raw content embeddings to new content embeddings. Triplet Selector is a training instance random selecting algorithm for reducing Triplet Loss. Triplet Loss is a specific loss function combining two distance-adjusting objects between various instances.

\subsubsection{Embedding Network}
\label{design:embeddingnet}
In \BotTriNet, the Embedding Network is a symmetrical multilayer perceptron for representing raw content embeddings. The input layer and output layer of the embedding network have the same number of dimensions. In our experiments, the simplest three-layer structure is good enough where the input and output layers are N-dimensional vectors, and the middle layer is 2N-dimensional. When we increased the count of layers and the increasing count of dimensions in the middle layers synchronously, the bot prediction accuracy improved. We concluded that a simple linear representation is enough to refine the raw embeddings effectively, and embedding networks with more complex structures could improve the result.

\subsubsection{Triplet Loss}
The Triplet Loss is a function that helps to find the best parameters of the Embedding Network, achieving two goals: minimizing the distance between an anchor and its positive example while maximizing the distance between the anchor and its negative example. At each iteration, the Triplet Network uses the gradient descent method to minimize the following loss function by adjusting the parameters of the Embedding Network: \\
\\
$Loss = \max(0,  \lVert f(x_a)-f(x_p) \rVert - \lVert f(x_a)-f(x_n) \rVert + m)$
\\
\\
with $x$ the raw embeddings for anchor, positive and negative embeddings, $f(x)$ the Embedding Network outputting represented embeddings, $\lVert \rVert$ a distance metric, and $m$ a hyperparameter to set the minimal distance between negative and positive embeddings. We set $m = 0.5$ and use the Euclidean as the distance metric.

\subsection{Triple Selector}
\label{design:selector}
The triplet learning process is the procedure of reducing \emph{Triplet Loss}. During the iteration, \BotTriNet randomly selects an \Triple, a tuple consisting of an anchor, a positive, and a negative instance, used to compute the triplet loss and backpropagate the gradients for optimizing the parameters of the embedding network. 

The Triplet Selector algorithm is essential to making the embedding network robust and accurate. We consider two problems that greatly influence bot detection accuracy: How to set the initial embedding of an anchor? How to select positive and negative samples properly?

\subsubsection{Centric Anchor Initialization}
\label{subsubsec:center} 
\BotTriNet uses the average raw content embeddings to embed an anchor of account rather than choosing arbitrary content embedding as the initial anchor embedding. We borrowed ideas from the iterative process of the clustering algorithm, using the mean value to represent the cluster's central semantics, expecting it to make the triplet network converge faster. We evaluate the benefit of the center initialization approach by comparing it with random initialization in Section~\ref{subsec:eval_selector}. The result shows that under the same iteration time, \emph{averaging initialization} performs better.

\subsubsection{Adapted Samples Selection}
\label{subsubsec:filter} 
In the original definition, \emph{positive samples} refer to images with the same class with an anchor, and  \emph{negative samples} have the same class. Similarly, \BotTriNet sets positive samples as content embeddings produced by the same account (anchor) and sets \emph{negative samples} as an arbitrary content embedding with the opposite label with the anchor. The negative content embedding must belong to another anchor. The setting of negative sample selection is trick but efficient to achieve two goals simultaneously:  maximizing the distances between bots and genuine users;  not affecting distances inside the same groups (bot account or genuine account). We evaluate the effect of the selection approach in Section~\ref{subsec:eval_selector}, which improves the performance compared to the original definition of negative sample selection.

\subsection{Detect Bots via Account Embeddings}
\BotTriNet uses \emph{account embeddings}, not isolated content embeddings, for bot detection. We define account embedding as profiling one account's long historical behaviors, containing all semantic content information in the past. We consider \emph{average pooling} is good enough in extensive research practice, and our focus is to explore the significant improvement of the triplet learning representation. Therefore, the system still applies \emph{average pooling} to product account embeddings based on content embeddings with the same account, forming same-dimension feature vectors.

The system integrates various classifiers for bot detection, e.g., Support Vector Machine, Multilayer Perceptron, and Random Forest. \BotTriNet uses the \emph{account embeddings} as the feature vectors and the labels of accounts in the training data to fit a classification model. It predicts the labels of prediction instances using the account embeddings through the same embedding generation and representation process.

\section{Evaluation}
\label{sec:evaluation}

\subsection{Ground-truth and Metrics}
\begin{table}[h]
\centering
\caption{Ground truths constructed by bot sets in \Cresci data set}
\label{tab:gt}
\vspace{-6pt}
\renewcommand{\arraystretch}{1.1}
\small
\begin{tabular}{|l|p{2.4in}|}
\hline
{\bf Ground-truth} & {\bf Description} \\
\hline
GT-SBOT1 &  the set of social bot 1, and genuine users \\
\hline
GT-SBOT2 &  the set of social bot 2, and genuine users \\
\hline
GT-SBOT3 &  the set of social bot 3, and genuine users \\
\hline
GT-SBOT &  the sets of social bot 1 to 3, and genuine users \\
\hline
GT-TBOT &  the set of traditional bot 1, and genuine users \\
\hline
GT-FBOT & the set of fake followers, and genuine users\\
\hline
GT-ABOT & all above bot sets, and genuine users\\
\hline
\end{tabular}
\end{table}

% Table generated by Excel2LaTeX from sheet 'baseline new'
\begin{table*}[htbp]
  \centering
  \caption{Accuracy and F1score performance on baseline approach and \BotTriNet}
  	\vspace{-6pt}
    \begin{tabular}{|c|c|c|c|c|c|c|c|c|c|}
    \hline
     & \multicolumn{3}{c|}{\textbf{Negative Instances}} & \multicolumn{3}{c|}{\textbf{Accuracy}} & \multicolumn{3}{c|}{\textbf{F1score}} \\
     \hline
\textbf{Groud Truth}& \textbf{$Count_a$} & \textbf{$Count_t$} & \textbf{$Count_m$} & \textbf{Baseline} & \textbf{BotTriNet} & \textbf{Gain} & \textbf{Baseline} & \textbf{BotTriNet} & \textbf{Gain} \\
    \hline
    \textbf{GT-SBOT1} & 991   & 1610034 & 1625  & \textcolor[rgb]{ 0,  .69,  .314}{99.36\%} & \textcolor[rgb]{ 0,  .69,  .314}{\textbf{99.51\%}} & 0.15\% & \textcolor[rgb]{ 0,  .69,  .314}{99.35\%} & \textcolor[rgb]{ 0,  .69,  .314}{\textbf{99.51\%}} & 0.16\% \\
    \hline
    \textbf{GT-SBOT2} & 3457  & 428542 & \textbf{124} & \textcolor[rgb]{ 1,  0,  0}{80.96\%} & \textcolor[rgb]{ 0,  .69,  .314}{\textbf{95.59\%}} & \textbf{14.63\%} & \textcolor[rgb]{ 1,  0,  0}{78.74\%} & \textcolor[rgb]{ 0,  .69,  .314}{\textbf{94.82\%}} & \textbf{16.08\%} \\
    \hline
    \textbf{GT-SBOT3} & 464   & 1418626 & 3057  & \textcolor[rgb]{ 0,  .69,  .314}{96.49\%} & \textcolor[rgb]{ 0,  .69,  .314}{\textbf{97.16\%}} & 0.67\% & \textcolor[rgb]{ 0,  .69,  .314}{95.59\%} & \textcolor[rgb]{ 0,  .69,  .314}{\textbf{96.46\%}} & 0.87\% \\
    \hline
    \textbf{GT-TBOT} & 1000  & 145094 & \textbf{145} & \textcolor[rgb]{ 1,  0,  0}{86.17\%} & \textcolor[rgb]{ 0,  .69,  .314}{\textbf{92.59\%}} & \textbf{6.42\%} & \textcolor[rgb]{ 1,  0,  0}{73.68\%} & \textcolor[rgb]{ 0,  .69,  .314}{\textbf{88.92\%}} & \textbf{15.24\%} \\
    \hline
    \textbf{GT-FBOT} & 3202  & 196027 & \textbf{61} & \textcolor[rgb]{ 1,  0,  0}{83.45\%} & \textcolor[rgb]{ 0,  .69,  .314}{\textbf{93.10\%}} & \textbf{9.65\%} & \textcolor[rgb]{ 1,  0,  0}{83.03\%} & \textcolor[rgb]{ 0,  .69,  .314}{\textbf{93.08\%}} & \textbf{10.05\%} \\
    \hline
    \textbf{GT-SBOT} & 4912  & 3457202 & \textbf{704} & \textcolor[rgb]{ 1,  0,  0}{83.12\%} & \textcolor[rgb]{ 0,  .69,  .314}{\textbf{96.44\%}} & \textbf{13.32\%} & \textcolor[rgb]{ 1,  0,  0}{75.77\%} & \textcolor[rgb]{ 0,  .69,  .314}{\textbf{94.74\%}} & \textbf{18.97\%} \\
    \hline
    \textbf{GT-ABOT} & 9114  & 3798323 & \textbf{417} & \textcolor[rgb]{ 1,  0,  0}{82.60\%} & \textcolor[rgb]{ 0,  .69,  .314}{\textbf{96.18\%}} & \textbf{13.58\%} & \textcolor[rgb]{ 1,  0,  0}{69.88\%} & \textcolor[rgb]{ 0,  .69,  .314}{\textbf{93.03\%}} & \textbf{23.15\%} \\
    \hline
    \end{tabular}%
  \label{tab:eval_bs}%
\end{table*}%

\emph{Ground-truth} plays an essential role in evaluating the prediction model's performance through various metrics. \Cresci has five groups of bot accounts, three of which are social bots. To evaluate the accuracy robustness of \BotTriNet on different bot account categories, we constructed seven ground-truth sets as shown in Table~\ref{tab:gt}, each consisting of the corresponding bot accounts and genuine users. Genuine users are set as negative instances (label = 0), and bot accounts are set as positive instances (label = 1).

We separated each ground truth into two parts by random splitting: a \emph{training set} (occupying 70 percent) and a \emph{testing set} (occupying 30 percent). According to the predicted label and actual label of each sample, the result turns to four situations: (i) True Positive (TP); (ii) False Positive(FP);  (iii) True Negative (TN); (iv) False Negative (FN). We applied two wide-used measures \emph{accuracy} and \emph{f1-score} to evaluate \BotTriNet detection performance. \textbf{Accuracy} equals $\frac{TP+TN}{TP+FP+TN+FN}$, reflecting the percent of correctly predicted instances, including all actual and negative instances in their actual situation. \textbf{F1-score} is a comprehensive result of \emph{precision} and \emph{recall}, equaling $\frac{2 \cdot precision \cdot recall}{precision+reall}$. \emph{Precision}, equaling $\frac{TP}{TP+FP}$, is the correctness of detected positive instances. The higher f1-score is,  both the precision and recall are higher.

\subsection{\BotTriNet v.s. Baseline Approach}
\label{subsec:eval_base}
We set a baseline approach to make contrasts to show the effectiveness of \BotTriNet on various ground truths and how \BotTriNet helps break the bottleneck of accuracy on content-less bot categories. In detail, it uses the same training set, testing set, and classifying algorithm (all Random Forest) but chooses raw content embeddings to generate account embeddings as features. 

Table ~\ref{tab:eval_bs} shows the accuracy and f1score for the baseline and \BotTriNet. Clearly, for \textbf{GT-SBOT1} and \textbf{GT-SBOT3}, just using the baseline approach has already shown a good performance, but other ground truths are not acceptable. For further investigation, we counted several statistical indicators for each ground truth. $Count_a$ is the number of negative instances (bots) in the ground truth, and $Count_t$ is the total number of tweets all bots send. (The count numbers differ from the original \Cresci published because we removed the accounts without tweeting logs.) We define $Count_m$ as the average count of tweets, equaling $\frac{Count_t}{Count_a}$. We observed a linear correlation between $Count_m$ and the prediction performance of the baseline approach. It shows that lower $Count_m$ achieves a lower performance. We infer that bot categories with less content and activities are naturally hard to detect.

However, \BotTriNet improved the accuracy and f1score of all ground truths (bot categories) to very high, with an average accuracy of 95.79\% and an average f1score of 94.37\%. In five ground truths where the baseline approach performed poorly, \BotTriNet gained an average accuracy improvement of 11.52\% and an average f1score improvement of  16.70\%. \BotTriNet also showed robustness when detecting with mixed multiply bot categories like \emph{GT-SBOT}, \emph{GT-ABOT}.

\begin{table}[htbp]
  \centering
  \caption{Performance comparison among \BotTriNet and various high-impact spambot detection systems}
  	\vspace{-6pt}
    \begin{tabular}{|l|r|r|r|r|}
    \hline
          & \multicolumn{2}{c|}{GT-SBOT1} & \multicolumn{2}{c|}{GT-SBOT3} \\
    \hline
    Approaches & \multicolumn{1}{l|}{Accuracy} & \multicolumn{1}{l|}{F1score} & \multicolumn{1}{l|}{Accuracy} & \multicolumn{1}{l|}{F1score} \\
    \hline
    BotOrNot? \cite{davis2016botornot} & 73.40\% & 26.10\% & 92.20\% & 76.10\% \\
    \hline
    Ahmed et al. \cite{ahmed2013generic} & 94.30\% & 94.40\% & 92.30\% & 92.30\% \\
    \hline
    Cresci et al. \cite{cresci2016dna} & 97.60\% & 97.70\% & 92.90\% & 92.30\% \\
    \hline
    Feng et al. \cite{wei2019twitter} & 96.10\% & 96.30\% & 92.90\% & 92.60\% \\
    \hline
    BotTriNet & \textbf{99.51\%} & \textbf{99.51\%} & \textbf{97.16\%} & \textbf{96.46\%} \\
    \hline
    \end{tabular}%
  \label{tab:eval_compare}%
\end{table}%

\subsection{Compare With Previous Bot Detection Systems}
We especially compared \BotTriNet with the most advanced content-based bot detection technique \cite{wei2019twitter} and its comparison objects. BotOrNot? \cite{davis2016botornot}, Ahmed et al. \cite{ahmed2013generic} and Cresci et al. \cite{cresci2016dna} are high-impact research works on bot detection, and Feng et al. \cite{wei2019twitter} made a Bidirectional LSTM approach to beat them by comparing accuracy and f1score on GT-SBOT1 and GT-SBOT3. We also compared the performance of GT-SBOT1 and GT-SBOT3, and Table ~\ref{tab:eval_compare} shows the result. Our approach outperformed all previous works on the above two ground truths.

\subsection{The Effect of two \Triple Selector Methods}
\label{subsec:eval_selector}
% Table generated by Excel2LaTeX from sheet 'selection new'
\begin{table}[htbp]
  \centering
  \caption{F1score under different \Triple Selector Settings}
  \vspace{-6pt}
    \begin{tabular}{|c|c|c|c|c|}
    \hline
    \textbf{Settings} & \textbf{GT-TBOT} & \textbf{GT-FBOT} & \textbf{GT-SBOT} & \textbf{GT-ABOT} \\
    \hline
    \textbf{Random Anchor} & 88.42\% & 90.72\% & 93.69\% & 91.90\% \\
    \hline
    \textbf{Old Negative} & 87.44\% & 91.31\% & 93.02\% & 92.60\% \\
    \hline
    \textbf{BotTriNet} & \textbf{88.92\%} & \textbf{93.08\%} & \textbf{94.74\%} & \textbf{93.03\%} \\
    \hline
    \end{tabular}%
  \label{tab:eval_select}%
\end{table}%

In Section~\ref{design:selector}, we proposed two approaches in \Triple Selection for social account modeling, respectively called \emph{Centric Anchor Initialization} and \emph{Adapted Samples Selection}. We did two comparing experiments under the same epoch time to prove that both will improve the detection accuracy compared to the baseline method. In Tabel~\ref{tab:eval_select}, the first baseline experiment uses randomly selected content embedding to embed an anchor (called Random Anchor). The second baseline experiment uses the traditional definition of Triplet Network, which means all negative samples come from other anchors (called Old Negative). Except for the two specific changes, all settings are the same with standard \BotTriNet. The result shows that without \emph{Centric Anchor Initialization} and \emph{Adapted Samples Selection}, the f1score drops, respectively, and the complete \BotTriNet performs the best.

\section{Conclusions}
We study the bot account detection problem on online social network platforms using a popular real-world dataset with abundant textual content posted by accounts. To investigate a general paradigm for improving accuracy drastically, We present \BotTriNet, a unified embedding and tuning system for bot detection, employing a triplet network to represent raw embeddings into more efficient values.

\BotTriNet is inspired by the concept of anchors in the face recognition field, where each anchor could have various image samples. Similarly, social accounts produce many contents with semantic information,  reflecting the account's personal style and expression habits, whether it is a genuine user or a bot. \BotTriNet designs a domain-specific Triple Selection Algorithm to model the correlation among social accounts and actively increase the distance between feature vectors of bots and genuine users.

Our system outperforms previous research with significant progress, particularly breaking the accuracy bottleneck of content-less bot sets, such as traditional bots and fake followers. It also demonstrates the robustness of high accuracy faced with multi-class datasets.

\bibliographystyle{IEEEtran}
\bibliography{paper}

\end{document}